\pdfoutput=1
\newif\ifextended
\extendedtrue

\documentclass[10pt,twocolumn]{article}

\usepackage[utf8]{inputenc}
\usepackage[T1]{fontenc}
\usepackage{lmodern}
\usepackage[margin=0.75in]{geometry}
\usepackage{booktabs}
\usepackage{array}
\usepackage{graphicx}
\usepackage{amsmath}
\usepackage{listings}
\usepackage{xcolor}
\PassOptionsToPackage{hyphens}{url}
\usepackage[hidelinks]{hyperref}
\usepackage{microtype}

\newcolumntype{L}[1]{>{\raggedright\arraybackslash}p{#1}}
\newcommand{\tabhead}[1]{\textbf{#1}}

\newcommand{\armS}{\textsc{schema}}
\newcommand{\armM}{\textsc{manual}}
\newcommand{\armH}{\textsc{hollow}}
\newcommand{\armC}{\textsc{contract}}
\newcommand{\mGLM}{GLM}
\newcommand{\mDS}{DeepSeek}
\newcommand{\mSON}{Sonnet~5}
\newcommand{\mGPT}{GPT-5.6}

\ifextended
  \newcommand{\appref}[1]{Appendix~\ref{#1}}
\else
  \newcommand{\appref}[1]{the extended version~\cite{extended}}
\fi

\title{\vspace{-2em}\bfseries Which Part of the Context Layer Does the Work?\\
\large Separating Semantic Content from Retrieval Scaffolding in Text-to-SQL Agents}

\author{
  Qing Ye\\
  \small Independent Researcher\\
  \small\texttt{qing.leafnode@gmail.com}
}
\date{}

\begin{document}
\raggedbottom
\maketitle

\begin{abstract}
\emph{Context layers}, curated documentation that an analytics agent
fetches at query time, produce large accuracy gains on text-to-SQL
benchmarks. A with/without comparison cannot say which part of
the layer does the work: the semantic content, the retrieval scaffolding that
delivers it, or the pre-computed views that usually accompany it.

We report a four-arm ablation on DABStep on four models that separates
the three. The instrument is a data contract: a YAML artifact that
carries a domain's semantics and the rules an agent's tools enforce. One arm
empties every field of prose in the frozen contract while holding the tool
surface, retrieval instruction, table allow-list and operation rules
byte-for-byte fixed. Compiling the contract's own SQL expressions into views gives the ceiling
a pre-computed layer would reach: gold on all 176 tasks it covers.

Content dominates. It raises hard-task accuracy from 13.9\% to 55.1\%, 22.6\%
to 56.6\%, 22.9\% to 68.4\% and 37.0\% to 77.4\%, beating the same knowledge
pasted into the prompt on every model. Scaffolding without content is worth 0 to 5 points on two flash models and
14 to 15 on two frontier models. Against the
compiled ceiling the contract arm's shortfall is a failure to derive, and it
falls from 39 points to 5 with model capability. The contract beats the
prompt because the rule it needs is one lookup away rather than buried in a
long prompt: its SQL carries the fee semantics up to 98\% of the time
against the prompt arm's 4\%, and at the lowest cost per correct answer on
three of four models. For practitioners: semantics first, scaffolding second, pre-computed macros only where an agent
demonstrably fails to derive. Ungoverned arms submitted 166 mutating
statements; governed arms none. The gain is confined to the contract's
domain. On one model the benchmark's own withheld golds grade the contract
arm at 51.9\% against 18.8\% for the prompt baseline.

\end{abstract}

\section{Introduction}
\label{sec:intro}

An analytics agent pointed at a warehouse it does not understand fails in a
characteristic way. It reads the schema, writes plausible SQL, and returns a
number that is wrong for a reason no column name records: a code whose values
mean something specific to the business, a join whose cardinality depends on a
convention, a \texttt{NULL} that means \emph{applies to everything} rather than
\emph{unknown}. The query runs. The answer is confidently wrong.

The industry's answer is the \emph{context layer}: curated documentation
stored alongside the data and fetched by the agent at query time. The reported
gains are large. MotherDuck's Guides raise DABStep accuracy by 72 percentage
points while cutting cost by roughly 55\%~\cite{motherduck-guides}, and
similar claims accompany every semantic-layer and context-layer product now
shipping.

These results share a shape: system with the layer against system without
it, two accuracy numbers, a large difference. That establishes \emph{that}
context layers work. It cannot establish \emph{which part} works, and the part
matters because the parts cost different amounts. A context layer as deployed
bundles at least three things:

\begin{enumerate}
\itemsep2pt
\item \textbf{Semantic content}: prose describing what the columns mean, how
      the metrics are defined and which conventions hold.
\item \textbf{Retrieval scaffolding}: the tools that fetch that content, the
      instruction to fetch it before writing SQL, and the narrowed table
      surface that comes with a curated catalogue.
\item \textbf{Pre-computed artifacts}: views and macros that encode a metric
      once, so the agent calls it instead of deriving it.
\end{enumerate}

Semantic content is written once per domain and often exists already, in a
vendor manual or a data dictionary. Scaffolding is engineering, paid once.
Pre-computed macros are paid \emph{per metric}, forever, and cover only the
metrics somebody anticipated. A practitioner deciding where to spend needs to
know which of the three bought the 72 points.

Prior work has isolated the first. BIRD withholds and supplies human-written
evidence, worth 20 points to GPT-4~\cite{bird}; KaggleDBQA reports that
database documentation doubles parser accuracy~\cite{kaggledbqa}; an ontology
layer over an enterprise schema moves GPT-4 from 16\% to
54\%~\cite{sequeda-kg}. None separates content from the scaffolding that
delivers it, and none says what pre-computing the same content would have
bought.

\subsection*{Contribution}

We report a controlled ablation on DABStep~\cite{dabstep}, run on four model
families, that separates all three. The instrument is a \emph{data contract}:
a YAML artifact that carries a domain's semantics (what a fee is, how a rule
matches a transaction) together with the rules an agent's tools enforce
(which tables may be read, which operations are forbidden). Two properties of
that artifact make the separation possible.

First, the contract can be \emph{hollowed}. A generator empties every field
of prose --- domain summaries, metric descriptions, the SQL expression that
carries the fee formula, column and relationship descriptions --- and leaves
names, types, structure, the table allow-list, the forbidden operations and
all nine governed tools byte-for-byte identical. An agent given the hollow
contract still receives the instruction to look things up, still has the
tool, and still calls it. It is simply told nothing when it does.

Second, the contract \emph{compiles}. Its declared SQL expressions, transcribed
into two views, reproduce the benchmark's gold answer on every one of the 176
tasks they cover. That is the accuracy a pre-computed layer built from the
same content would reach, and it turns each arm's accuracy into a recovery
rate against a proven ceiling.

Together the two place the arms on a ladder: schema alone, scaffolding
without content, content delivered through scaffolding, and content
compiled. Beside the ladder sits the practitioner's baseline, the same
knowledge pasted verbatim into the prompt: content without scaffolding.
Reading the ladder gives four findings and one non-finding.

\textbf{Content dominates; scaffolding is real but secondary}
(Section~\ref{sec:decomposition}). Restoring the prose to the hollow contract
raises hard-task accuracy by 26 to 36 points on every model. Adding the empty
scaffolding to a bare schema is worth 0 to 5 points on the two flash models
(not significant over all tasks) and 14 to 15 on the two frontier models. The content step
exceeds the scaffolding step by $1.8{\times}$ to $6.6{\times}$, and on one
model is the whole effect.

\textbf{The contract beats the prompt because its rule is reachable, not
because it knows more} (Section~\ref{sec:why}). Both the contract arm and the
prompt arm are given the fee-matching rule; only the contract arm applies it.
Read in the submitted SQL, the contract arm writes every one of the six fee
clauses a task needs 39\%, 65\%, 55\% and 98\% of the time; the prompt arm,
3\%, 9\%, 4\% and 4\%. On one model the 24{,}000-character prompt is
indistinguishable from a bare schema on hard tasks, while the contract
carrying the same knowledge sits 45 points above both. The contract arm also
does the least work on every model, and is the cheapest way to buy a correct
answer on three of the four.

\textbf{The derivation gap closes with capability; the composition gap does
not} (Section~\ref{sec:ceiling}). On the 176 tasks the compiled contract
answers outright, the contract arm falls short of the ceiling by 39.2, 37.3,
26.1 and 5.1 points in order of model capability. On tasks that compose the
same semantics further, the shortfall moves between 54 and 40 points with no
clean ordering. The cost of making an agent derive a query rather than call a
macro is transient; the cost of the reasoning built on top of it is not yet.

\textbf{Declared rules deter mutation without being enforced}
(Section~\ref{sec:governance}). Ungoverned arms submitted 166 mutating
statements across 26 tasks and corrupted the warehouse twice; governed arms
submitted none and never attempted one, so the validator behind the rules
never fired.

The non-finding: on the 38, 30, 38 and 38 hard tasks outside the domain the
contract describes, it holds no detectable advantage over the prompt, and on
the strongest model the bare schema leads that slice. The gain is
domain-specific by construction. That is what a domain contract should do,
and it is the sharpest limit on what one benchmark can show.

\textbf{What a practitioner should take from this.} Spend on the
semantics first: write down what the columns mean and how the metrics are
defined, once per domain, in a form an agent can fetch one rule at a time.
That step bought most of the gain on every model, and the prose often
exists already. Build the retrieval scaffolding second; on frontier models
it is worth 14 to 15 points, on flash models close to nothing. Pre-compute
a macro only for a metric the agent demonstrably fails to derive, because
the shortfall macros close is already 5 points on the strongest model and
falling with capability, while the per-metric cost of writing and
maintaining them falls with nothing.

\section{Background and Related Work}
\label{sec:background}

\subsection{DABStep}
\label{sec:dabstep}

DABStep~\cite{dabstep} is a data-analysis benchmark built by Adyen and Hugging
Face around a payments warehouse: 138{,}236 transactions, a fee-rule table,
merchant and merchant-category-code (MCC) reference data, and two documents
of human-written context, \texttt{manual.md} (22{,}127 characters) and a
dataset README. Three of its properties shape this study.

\textbf{Its difficulty is semantic, not syntactic.} The hard split does not
require exotic SQL. It requires knowing the fee-matching rule: a \texttt{NULL}
in a \texttt{fees} column means \emph{this rule applies to every value of that
column}, not \emph{unknown}, and matching additionally involves list
membership over account type, authorisation characteristics indicator (ACI)
and merchant category code. An agent that reads the schema and writes
idiomatic SQL gets a clean, wrong number, which is the failure a context
layer claims to fix. Among text-to-SQL benchmarks this sets DABStep apart
from Spider~\cite{spider} and Spider~2.0~\cite{spider2}: its difficulty is
documentation, not schema scale.

\textbf{Golds are withheld and grading is centralised}, which raises the cost
of the overfitting that makes many text-to-SQL results hard to
trust~\cite{ladder,leaderboard-illusion}. It does not remove it: the sources
an agent reads are public, and Section~\ref{sec:golds} recovers 59
hard-split answers from them.

\textbf{Its question space is small.} DABStep expands 450 test tasks from a
smaller set of core questions by varying merchants, schemes, months and
thresholds, following GSM-Symbolic's parameterisation~\cite{gsm-symbolic}.
Normalising those varying entities away leaves 113 distinct question shapes,
of which 12 cover half the benchmark and 27 cover three quarters. In our task
set 294 of 332 hard tasks concern fees. Section~\ref{sec:threats} treats this
as the sharpest limit on the result, and Section~\ref{sec:prior} explains
why it favours pre-computed views.

The benchmark's own paper reports its strongest agent at 14.55\% on the
hard split against 76.39\% on the easy one~\cite{dabstep}. Hard-split
accuracies from credible named leaderboard entries since cluster around
20--26\% (Google's \texttt{simple\_baseline} at 26\%, an Adyen GPT-5.4 entry
at 22.2\%, a Hugging Face Claude~4 Sonnet entry at 19.8\%), and we use that
band as the validity check on our reimplementation of the prompt-based
approach (Section~\ref{sec:headline}).

\subsection{Three kinds of layer, and when their semantics are compiled}
\label{sec:layers}

Three families of approach exist and are routinely conflated. \emph{Semantic
layers} in the classical sense --- dbt metrics, Cube, LookML --- define
metrics as executable artifacts, so the agent calls a metric rather than
deriving it and correctness is inherited from the definition. \emph{Context
layers} store natural-language documentation next to the data and retrieve
it at query time; MotherDuck Guides~\cite{motherduck-guides} is the
best-documented instance. The agent still writes the SQL, but writes it
knowing what the columns mean. \emph{Data contracts} originate in governance
rather than assistance: a declarative artifact stating what a dataset
contains, who owns it and what may be done with it, with the Open Data
Contract Standard~\cite{odcs} the emerging specification. They inherit a
longer line of dataset documentation, from datasheets~\cite{datasheets} to
Croissant's metadata a tool loads~\cite{croissant}. MotherDuck's own primer
draws the first line the same way: a semantic layer holds definitions
``written as logic that compiles to SQL'' and a context layer ``holds
everything else''~\cite{motherduck-primer}. The database community's framing
of the wider problem is TAG~\cite{tag}.

The artifact under test is of the third kind and serves both purposes at
once: the same YAML supplies the semantics the agent reads and the rules the
tool layer enforces (Section~\ref{sec:system}). The distinction that matters
for this paper is \emph{when} the semantics are compiled
(Table~\ref{tab:compile}). A macro is compiled ahead of time by a human and
covers the metrics somebody anticipated; a contract is compiled per query by
the model and covers whatever question arrives.

\begin{table}[t]
\centering
\small
\begin{tabular}{@{}L{0.27\columnwidth}L{0.26\columnwidth}L{0.33\columnwidth}@{}}
\toprule
 & \tabhead{Views / macros} & \tabhead{Declarative contract} \\
\midrule
Compiled & ahead of time, by a human & per query, by the model \\
Covers & anticipated metrics & any question in the domain \\
Cost & one macro per metric, maintained & one description per domain \\
Fails when & nobody anticipated the question & the model misapplies the rule \\
\bottomrule
\end{tabular}
\caption{Two ways to deliver the same semantics. The compiled contract of
Section~\ref{sec:ceiling} is the left column built from the right column's
content, which is what lets this paper measure the difference between them.}
\label{tab:compile}
\end{table}

\subsection{What has been measured, and what this paper adds}
\label{sec:prior}

\textbf{Context layers on DABStep.} MotherDuck's Guides --- markdown context
stored in the warehouse and fetched through an MCP server --- raise DABStep
accuracy by 72 points and cut cost per run by 55\%, with Gemini~3 Flash
answering 418 of 419 held-out questions (99.8\%)~\cite{motherduck-guides}; a
companion report reaches 98.6\% with a locally hosted 27B model at 4-bit
quantization~\cite{motherduck-local}. A second MotherDuck report makes the
contribution of pre-computation legible because it reports a progression
rather than a number~\cite{motherduck-semantic}: retrieving context fragments
by vector search ``capped out around 88\%''; baking the knowledge into the
warehouse as schema comments, macros and derived tables reached 93\%; a
hierarchical semantic layer over the raw data, authored by a large model and
refined iteratively, reached 100\%. Each step moves work out of the agent and
into an artifact prepared in advance. The system atop the validated leaderboard, at 89.95\% hard, is built the
same way, by validating candidate helper functions against the benchmark's
gold answers~\cite{nvidia-kgmon}; nothing in the benchmark forbids that,
and we allege no violation. The second-placed system, at 87.57\%, is
described in a post we could not retrieve in full~\cite{oceanbase-datapilot}.

A reader will otherwise stop at the difference between those numbers and our
best hard-split figure of 77.4\%, so we say now what it is made of. Pre-built
views are maximally effective when the question space is enumerable in
advance, which is this benchmark's construction (Section~\ref{sec:dabstep}).
A macro contributes nothing to a question nobody wrote it for; the
declarative layer measured here is the layer that must cover the
unanticipated question. \textbf{Our hard-split figures measure a harder task,
not a worse attempt at the same one}, and Section~\ref{sec:ceiling} puts a
number on how much harder: the compiled form of our own contract answers
53\% of the hard split outright and none of the rest. The two layers are
complementary. Macros optimise the head of the question distribution, a
contract covers the tail, and in a system with both, the contract is what
tells the agent which macro exists and when to call it. We do not claim the
contract approach scales better, only that it degrades differently: a
missing macro degrades to nothing, a contract to a model reasoning from a
description.

The tuned artifacts also do not travel: the author of the 100\% result
writes that ``the tuned artifact is coupled'' to the serving
model~\cite{motherduck-semantic}, and independent commentary on Guides urges
measuring ``whether the Guide generalizes to new questions, not whether it
clears the benchmark you tuned it against''~\cite{corrdyn-guides}. Our
contract was authored from the vendor manual alone, frozen before any
question was read, and run unchanged against four model families, which is
why it can be published with the paper (\appref{app:leaderboard}).

\textbf{Related methods.} Emptying a component's content while holding
its form fixed, to see what the form alone is worth, is also the move
of~\cite{min-demonstrations}, applied there to demonstrations and here to a
whole retrieval layer. Two results bear on the delivery mechanism of
Section~\ref{sec:why}: retrieval quality falls for material buried
mid-prompt~\cite{lost-in-the-middle}, and a tool's documentation shapes
what an agent does with it~\cite{tool-documentation}. On governance, CaMeL
enforces capabilities outside the model~\cite{camel}, $\tau$-bench
measures compliance with a prompt-stated policy~\cite{tau-bench}, and RuLES
finds models breaking simple stated rules with no adversary
present~\cite{rules}, the expectation Section~\ref{sec:governance} runs
against. Reporting cost beside accuracy follows~\cite{agents-that-matter};
paired testing across several models is
standard~\cite{semantic-paired,dbt-benchmark}, and McNemar's test for
learners executed once is Dietterich's recommendation~\cite{dietterich}.

What is new here is the decomposition, not the accuracy: four arms rather
than with/without, a control that removes content while holding the
scaffolding byte-for-byte fixed, a ceiling compiled from the same content,
an artifact frozen and digest-pinned before any question was read, and
6{,}416 transcripts released.

\section{The System Under Test}
\label{sec:system}

\texttt{agentic-data-contracts} is an open-source Python library that loads a
data contract and exposes a warehouse to an agent through nine governed
tools. The library is the instrument, not the contribution, and the
experiment is designed so that its identity does not matter: the hollow arm
runs the identical library, tools and rules with the prose removed.

\subsection{The contract}
\label{sec:contract}

The contract is two YAML files, 60{,}181 characters together.
\texttt{contract.yml} declares the five tables the agent may read, the eleven
operations it may not issue (\texttt{DELETE}, \texttt{DROP},
\texttt{UPDATE}, \texttt{INSERT}, \texttt{CREATE}, \texttt{ALTER} and five
more), and eight domains, each with a one-line summary and a prose
description; the \texttt{fees} description restates the manual's account of
every column of the fee-rule table. \texttt{semantic.yml} declares 14
metrics, each with a description and an executable
\texttt{sql\_expression}, the five tables with a description per column, and
one relationship. Figure~\ref{fig:contract} shows one metric as it appears in
the contract and in its hollow counterpart. Seven metric descriptions carry a
sentence beginning \texttt{INTERPRETATION}, marking where the contract
commits to a reading the manual does not force; Section~\ref{sec:threats}
returns to them.

The same file supplies both the semantics the agent reads and the rules the
tool layer enforces. The agent reads the first through
\texttt{lookup\_domain} and \texttt{lookup\_metric}; the tool layer checks
every statement against the second before it reaches the database.

\begin{figure*}[t]
\begin{minipage}[t]{0.63\textwidth}
\begin{lstlisting}
- name: fee_rule_matches_transaction
  description: >
    True when fee rule `f` applies to transaction `p` of merchant `m`, on
    every field that a single transaction can decide. The manual's rule is
    that "if a field is set to null it means that it applies to all possible
    values of that field", so each clause below is satisfied either by the
    field being null or by the field agreeing with the transaction. [...]
    INTERPRETATION: the list-typed fields (account_type, aci,
    merchant_category_code) are never null in the annexed `fees` data --
    they express "applies to all values" as an empty list -- so each clause
    accepts an empty list as the wildcard the manual describes.
  sql_expression: >
    (f.card_scheme IS NULL OR f.card_scheme = p.card_scheme)
    AND (f.is_credit IS NULL OR f.is_credit = p.is_credit)
    AND (f.aci IS NULL OR len(f.aci) = 0 OR list_contains(f.aci, p.aci))
    AND (f.intracountry IS NULL
         OR (f.intracountry = 1) = (p.issuing_country = p.acquirer_country))
    [... three more clauses: account_type, merchant_category_code,
     capture_delay mapped from days to the manual's bands]
  source_model: main.fees
  domains: [fees]
\end{lstlisting}
\end{minipage}\hfill
\begin{minipage}[t]{0.35\textwidth}
\begin{lstlisting}
- name: fee_rule_matches_transaction
  description: No definition is
    available for this metric.
  sql_expression: ''
  source_model: main.fees
  domains:
  - fees
\end{lstlisting}
\vspace{2pt}
{\small The same metric in the hollow contract. Name, source table and
domain membership survive; the description and the SQL expression do not.
Domain prose and column descriptions are emptied the same way.}
\end{minipage}
\caption{One of the 14 metrics of the frozen contract (left, abridged) and
its hollow counterpart (right). The contract arm can fetch the left with one
\texttt{lookup\_metric} call; the hollow arm fetches the right.}
\label{fig:contract}
\end{figure*}

\subsection{Nine tools and two-layer validation}
\label{sec:tools}

{\sloppy The tools fall into three groups. Four discover structure:
\texttt{describe\_table}, \texttt{preview\_table},
\texttt{lookup\_relationships} and \texttt{list\_metrics}. Three deliver
semantics: \texttt{lookup\_domain} returns a domain's prose,
\texttt{lookup\_metric} a metric's description and SQL expression, and
\texttt{trace\_metric\_impacts} the metrics that depend on a table. Two
handle queries: \texttt{inspect\_query} validates a statement and returns
either a pass or a reason, and \texttt{run\_query} validates and executes
it. Validation is two-layer. Every statement is parsed with
sqlglot~\cite{sqlglot} and checked statically against the table allow-list
and the forbidden operations; where a database adapter is available, a dry
run follows. A rejected statement returns its reason to the model rather
than an error to the user. The tool descriptions themselves carry 3{,}042
characters of procedural text (\texttt{inspect\_query} tells the model to
call \texttt{lookup\_metric} first), and that text is part of what the
hollow arm holds fixed.\par}

\subsection{The hollow contract}
\label{sec:hollow}

The hollow contract is generated from the frozen contract by a script, not
written by hand, so the only difference between the two artifacts is the one
the generator makes, and it is auditable as a diff. Emptied: domain summaries
and descriptions, metric descriptions, every \texttt{sql\_expression}
(the fee formula lives there), and all column and relationship descriptions.
Retained byte-for-byte: every name, every type, the structure, the table
allow-list, the forbidden operations and the identical nine-tool surface. An
agent that calls \texttt{lookup\_domain} against it is told \emph{``No
documentation is available for this domain.''}

The control is verified in both directions. The test suite asserts that no
6-gram of \texttt{manual.md} survives into the hollow contract, and that the
real contract does share 6-grams with it; the second half is what makes the
first meaningful. The transcripts confirm that the retrieval loop ran: on
\mGLM{} the hollow arm called \texttt{lookup\_domain} on 272 of 401 tasks and
received the placeholder 282 times. Prompt length is deliberately not held
constant (1{,}984 characters against the contract arm's 2{,}475): padding the
hollow arm with text that reads as content but carries none would be a
different treatment, not a cleaner control. The variable under test is
knowledge, not tokens.

\subsection{Provenance and freeze}
\label{sec:freeze}

The most obvious way to fake this result is to write a contract that answers
the benchmark. The contract's header records that every fact in it came from
\texttt{manual.md} and the dataset README and nothing else, which a reader
can verify by reading the 60{,}181-character contract against the
22{,}127-character manual. The artifact is content-hashed
(\texttt{sha256:e438ecf7\ldots}), was frozen roughly 35 commits before the
first run, and every result row records its digest, so a post-hoc edit would
invalidate every row.\footnote{On three of the four models the hollow arm's
rows carry the real contract's hash rather than the hollow artifact's, a
stamping defect that changes no result (\appref{app:harness}).} What cannot be verified from outside is that
no benchmark question was read while the contract was authored. We claim
provenance; we cannot prove intent.

\section{Experimental Design}
\label{sec:design}

\subsection{Two factors, four arms}
\label{sec:arms}

Every arm sees the same tasks, the same warehouse, the same model, the same
answer-format instruction and the same scorer. The arms vary two factors
(Table~\ref{tab:arms}): whether the agent has the governed tools and the
instruction to consult them before writing SQL (\emph{scaffolding}), and
whether it is given the domain's semantics (\emph{content}).

\begin{table}[t]
\centering
\footnotesize
\setlength{\tabcolsep}{5pt}
\begin{tabular}{@{}lll@{}}
\toprule
 & no scaffolding & scaffolding \\
\midrule
no content & \armS{} & \armH{} \\
content    & \armM{} & \armC{} \\
\bottomrule
\end{tabular}
\caption{The four arms as a two-factor design (artifact names:
\texttt{schema\_only}, \texttt{contract\_hollow}, \texttt{manual\_prompt},
\texttt{contract}). \armH{} differs from \armC{} only in that its prose
fields are empty. \armM{} carries the same knowledge as \armC{} in its
original form, pasted into the prompt.}
\label{tab:arms}
\end{table}

\armS{} is the floor: three ungoverned tools (\texttt{list\_tables},
\texttt{describe\_table}, \texttt{execute\_sql}), a 281-character system
prompt, no documentation. It measures what the model can do from the schema
alone, and it doubles as our operational definition of model capability.

\armM{} is the published approach reimplemented. DABStep's own
\texttt{manual.md} is pasted verbatim into the system prompt, bringing it to
24{,}177 characters; the tools are identical to \armS{}. It validates the
harness against the leaderboard band, and it is the comparison a
practitioner actually faces: the alternative to a contract is not ignorance
but a long prompt.

\armC{} is the treatment: the nine governed tools over the frozen contract,
and a 2{,}475-character system prompt that names the domains and metrics
available and instructs the agent to look them up before writing SQL.

\armH{} is the control that makes the ablation an ablation: the same tools
and the same instruction over the hollow contract, with a 1{,}984-character
prompt. \armC{} differs from the baselines in tools, in a procedural
instruction and in content at once; \armH{} holds the first two fixed and
removes only the third.

Two qualifications on reading Table~\ref{tab:arms} as a factorial. The
content factor is the same knowledge in two forms, the manual itself and a
contract authored from the manual alone, so the \armM{}--\armC{} contrast
measures the form of delivery as well as its presence. And the scaffolding
factor changes the tool surface as well as the instruction, because that is
the bundle a curated catalogue actually brings; we measure the bundle, not
its parts.

\subsection{Four models}
\label{sec:models}

\begin{table}[t]
\centering
\footnotesize
\setlength{\tabcolsep}{3pt}
\begin{tabular}{@{}llllr@{}}
\toprule
\tabhead{Model} & \tabhead{Pinned id} & \tabhead{Route} & \tabhead{T} & \tabhead{Tasks} \\
\midrule
\mGLM{} & \texttt{glm-5.3-flash} & OpenRouter, fp8 & 0 & 401 \\
\mDS{} & \texttt{deepseek-v4-flash} & OpenRouter, fp8 & 0 & 279 \\
\mSON{} & Claude Sonnet~5 & Bedrock via gateway & --- & 401 \\
\mGPT{} & \texttt{gpt-5.6-sol} & OpenRouter & --- & 401 \\
\bottomrule
\end{tabular}
\caption{The four models, in the order used throughout: bare-schema hard
accuracy. This is not the order they were run in; \mSON{} ran last. Each is pinned to one provider endpoint with fallbacks disabled.
T is temperature; a dash means the route does not accept the parameter. \mDS{}'s
task count is its complete-case set, defined below.}
\label{tab:models}
\end{table}

We ran the full design once on each of four models
(Table~\ref{tab:models}), each pinned to a single provider endpoint with
fallbacks disabled so that quantization, price and throughput are fixed
within a run. We call \mGLM{} and \mDS{} the two flash models and \mSON{}
and \mGPT{} the two frontier models; that is the vendors' tiering, not a
measured quantity. Models appear throughout in order of bare-schema hard
accuracy (13.9\%, 22.6\%, 22.9\%, 37.0\%), the only capability axis declared
before any result was read; Section~\ref{sec:threats} records where that
axis disagrees with the contract arm's ordering. That is not the order the
runs were made in: \mSON{} was run last, after \mGPT{}, and sits third.
Every list of four numbers in this paper follows the table's order, \mGLM{},
\mDS{}, \mSON{}, \mGPT{}. The four runs are separate
experiments, not replicates: they differ in model, in provider and in one
library fix (\appref{app:harness}).

Reasoning effort is \texttt{medium} throughout, a nominal setting rather
than an equated one: under it the contract arm reasons for about 2{,}050
tokens per task on \mSON{} and about 350 on \mGPT{}. Temperature is 0 on
the two flash models; \mSON{}'s gateway rejects both \texttt{temperature}
and \texttt{seed} and \mGPT{} does not accept the parameter, so neither
frontier run has a determinism control. \mSON{} is reached through an
enterprise gateway on Anthropic's Messages API with cache control, because
the gateway's OpenAI-compatible route bills every input token fresh at a
multiple that differs by arm, a caching confound.

Two departures from the pre-registered design are recorded here. The
pre-registered primary model,
\texttt{deepseek-v4-pro}, was never swept: a 50-task probe of the
bare-schema arm showed it does not differ from the flash tier (27.5\%
against 25.0\%, $p{=}1.0$), so it would not have supplied the capability
contrast it was chosen for; \mGPT{} was substituted by the same criterion,
\mSON{} was added last, after \mGPT{} had run, and every contrast involving
them is exploratory
(\appref{app:harness}).
And the \mDS{} run lost 29\% of its rows to provider rate limiting,
near-uniformly across arms and with 114 of the 122 lost tasks losing all
four arms together, so we report it on the 279 tasks scoreable in all four
arms (its complete-case set), with intervals about 20\% wider
(\appref{app:truncation}).

\subsection{What is held fixed}
\label{sec:fixed}

Within a run everything else is fixed across the four arms: row limit (50
rows per result), tool-call cap (40), token budget, temperature, reasoning
effort, model, endpoint, quantization, task set, gold set, scorer and the
verbatim answer-format instruction. When an agent exhausts its output
budget the harness forces a final answer rather than discarding the row;
the rows so rescued fell mostly on the ungoverned arms and never on \armC{}
(on \mGLM{}, 41 in \armS{}, 18 in \armM{}, none in either governed arm), so
the policy works against the treatment. Three harness properties treated the arms unequally
(\appref{app:harness}): a validator defect on \mGLM{} that rejected
\texttt{COUNT(*)} as if it were \texttt{SELECT *} in the governed arms only,
so \mGLM{}'s contract figures are a floor on that axis; a 100-row preview
cap that reached the governed arms on 31 calls, in the treatment's favour;
and different provider routes across models, which makes cross-run cost
differences suggestive only. The validator defect, the preview cap and the
digest defect recorded there are fixed in the released harness; the rows
reported here are left as they ran.

\subsection{Tasks, golds and grading}
\label{sec:golds}

DABStep withholds golds for its 450 test tasks. We reconstruct them from the
leaderboard's published per-task files by plurality over answers the
benchmark's own grader marked correct, with a 75\% plurality threshold
(median realised share 0.981), which yields golds for 401 of 450 tasks (332
hard, 69 easy) after excluding five consensus answers that manual
verification showed to be wrong.
The three largest question families, 59 tasks, were recomputed directly from
the database using the manual's matching rule and reproduce exactly
(\appref{app:golds}).

The reconstruction was then measured rather than argued. Both arms of the
main contrast, \armC{} and \armM{} on \mGLM{}, were swept over all 450 tasks
and submitted to the leaderboard, which graded them against the withheld
golds: 51.9\% against 18.8\% on the 378 hard tasks, paired McNemar
$p{=}4.9\times10^{-28}$. On this model the main contrast is therefore graded
by the benchmark, not by us. On the 401 shared tasks the two gradings agree
on 382 (95.3\%); all 19 disagreements are wrong values our golds concealed,
all lie on the hard split, and all run the same way, so our hard-split
figures run about 5.7 points high for \armC{} and 3.0 for \armM{}. The
excluded tasks score like the kept ones under official grading, and a proxy
calibrated on these submissions bounds the leniency in every other arm and
model at under 3 points on any contrast (\appref{app:leniency}).

Answers are graded by DABStep's own scorer, vendored verbatim at a pinned
revision; our own earlier normaliser was stricter than the benchmark, and on
a 12-task pilot it manufactured a significant result in the contract's
favour that the official rules do not support.
Every arm sees every task, so all comparisons are paired and use McNemar's
exact test~\cite{mcnemar,dietterich}, reported with the discordant-pair
count and its split; proportions carry Wilson
intervals~\cite{wilson,brown-cai-dasgupta}; no multiplicity correction is
applied, and Section~\ref{sec:threats} says which results that exposes.
Tables report scored accuracy, correct over correct plus incorrect with
harness failures excluded; the released results carry strict accuracy, which
counts every non-correct row as wrong, beside it.

\section{Results: The Decomposition}
\label{sec:results}

\subsection{Accuracy and cost on four models}
\label{sec:headline}

\begin{table*}[t]
\centering
\small
\begin{tabular}{@{}lrrrr@{}}
\toprule
\tabhead{Arm} & \tabhead{Overall} & \tabhead{Hard} & \tabhead{Easy} & \tabhead{Cost} \\
\midrule
\multicolumn{5}{@{}l}{\emph{\mGLM{} (\texttt{glm-5.3-flash}, 401 tasks)}}\\
\armS{} & 22.7\% & 13.9\% & 65.2\% & \$0.76 \\
\armH{} & 26.4\% & 19.3\% & 60.9\% & \$0.85 \\
\armM{} & 31.7\% & 22.9\% & 73.9\% & \$0.96 \\
\armC{} & \textbf{57.9\%} & \textbf{55.1\%} & 71.0\% & \textbf{\$0.67} \\
\midrule
\multicolumn{5}{@{}l}{\emph{\mDS{} (\texttt{deepseek-v4-flash}, 279 tasks)}}\\
\armS{} & 33.0\% & 22.6\% & 77.4\% & \$1.79 \\
\armH{} & 31.5\% & 22.6\% & 69.8\% & \$1.52 \\
\armM{} & 52.0\% & 42.9\% & 90.6\% & \$1.69 \\
\armC{} & \textbf{61.3\%} & \textbf{56.6\%} & 81.1\% & \textbf{\$1.05} \\
\midrule
\multicolumn{5}{@{}l}{\emph{\mSON{} (Claude Sonnet~5, 401 tasks)}}\\
\armS{} & 31.9\% & 22.9\% & 75.4\% & \textbf{\$36.80} \\
\armH{} & 44.4\% & 38.0\% & 75.4\% & \$46.08 \\
\armM{} & 35.2\% & 23.8\% & 89.9\% & \$43.89 \\
\armC{} & \textbf{69.6\%} & \textbf{68.4\%} & 75.4\% & \$37.97 \\
\midrule
\multicolumn{5}{@{}l}{\emph{\mGPT{} (\texttt{gpt-5.6-sol}, 401 tasks)}}\\
\armS{} & 43.9\% & 37.0\% & 76.8\% & \$14.52 \\
\armH{} & 55.6\% & 51.2\% & 76.8\% & \$22.09 \\
\armM{} & 57.9\% & 50.3\% & 94.2\% & \textbf{\$14.42} \\
\armC{} & \textbf{78.8\%} & \textbf{77.4\%} & 85.5\% & \$21.32 \\
\bottomrule
\end{tabular}
\caption{Accuracy and cost, models in order of bare-schema capability. Hard
splits are $n{=}332$, 226, 332 and 332; easy are 69, 53, 69 and 69. \armC{}
is the most accurate arm on all four models, the cheapest on the two flash
models, and within about 3\% of the cheapest on \mSON{}. Arms are in ladder
order, so the two content-free arms are adjacent.}
\label{tab:headline}
\end{table*}

\begin{table*}[t]
\centering
\small
\setlength{\tabcolsep}{4pt}
\begin{tabular}{@{}lrrrr@{}}
\toprule
\tabhead{Comparison} & \mGLM{} & \mDS{} & \mSON{} & \mGPT{} \\
\midrule
\armS{} vs \armC{} & $3{\times}10^{-31}$ \,\, 14/\textbf{155} & $2{\times}10^{-15}$ \,\, 14/\textbf{93} & $1{\times}10^{-35}$ \,\, 11/\textbf{162} & $5{\times}10^{-32}$ \,\, 12/\textbf{152} \\
\armM{} vs \armC{} & $2{\times}10^{-17}$ \,\, 29/\textbf{134} & \textbf{0.0067} \,\, 30/\textbf{56} & $5{\times}10^{-27}$ \,\, 22/\textbf{160} & $2{\times}10^{-17}$ \,\, 12/\textbf{96} \\
\armH{} vs \armC{} & $9{\times}10^{-24}$ \,\, 23/\textbf{149} & $8{\times}10^{-17}$ \,\, 13/\textbf{96} & $2{\times}10^{-17}$ \,\, 25/\textbf{126} & $2{\times}10^{-17}$ \,\, 18/\textbf{111} \\
\armS{} vs \armM{} & $7{\times}10^{-6}$ \,\, 14/50 & $2{\times}10^{-12}$ \,\, 5/58 & \textbf{0.066} \,\, 15/28 & $2{\times}10^{-8}$ \,\, 22/78 \\
\armM{} vs \armH{} & 0.033 \,\, 55/34 & $4{\times}10^{-13}$ \,\, 63/6 & $1{\times}10^{-4}$ \,\, 27/\textbf{64} & 0.44 \,\, 58/49 \\
\armS{} vs \armH{} & \textbf{0.058} \,\, 20/35 & \textbf{0.61} \,\, 19/15 & $\mathbf{9{\times}10^{-9}}$ \,\, 14/\textbf{64} & $\mathbf{4{\times}10^{-7}}$ \,\, 20/\textbf{67} \\
\bottomrule
\end{tabular}
\caption{Every pairwise paired McNemar over all tasks: $p$-value, then
discordant pairs won by the left arm / by the right arm. The last row is the
scaffolding step of the decomposition, on which the flash and frontier
models split two and two. \mSON{}'s fourth row is its own result: the pasted
manual is not distinguishable from a bare schema.}
\label{tab:mcnemar}
\end{table*}

\armC{} wins every comparison over all tasks on all four models
(Tables~\ref{tab:headline} and~\ref{tab:mcnemar}), by 32.2, 13.7, 44.6 and
27.1 hard-task points over \armM{}. That margin is not monotone in
capability, and four points cannot support a story about why
(\appref{app:interaction}). On the two flash models \armC{} is also
the cheapest arm outright; on \mSON{} it is within about 3\% of the
cheapest; on \mGPT{} the \armM{} arm is cheaper, and
Section~\ref{sec:efficiency} says why.

Two checks tie the harness to the leaderboard. \armM{} reproduces the
published band where a comparison exists: 22.9\% hard on \mGLM{} against
the 20--26\% band for named entrants (18.8\% graded officially), and 23.8\%
on \mSON{} against the leaderboard's 19.8\% for Claude~4 Sonnet under the
same approach; on \mDS{} and \mGPT{} the same arm reaches 42.9\% and
50.3\%, above the band, which is a statement about those models. The check
is looser than it looks and in a direction that favours us, because our
grading runs 3 to 6 points high on the hard split (Section~\ref{sec:golds},
\appref{app:leniency}). The easy split is the negative check: \armC{} is not
the best arm on easy tasks on any model, and \armM{} leads on all four
(73.9\%, 90.6\%, 89.9\% and 94.2\% against 71.0\%, 81.1\%, 75.4\% and
85.5\%). The contract does nothing for questions that need no domain
knowledge, which is what it should do; an instrument that showed a contract
effect there would be measuring itself. On \mSON{} the lead is an
interaction rather than noise (62 of 69 against exactly 52 for each other
arm), and none of it is fee semantics: four of the 13 tasks involved have
their answer printed verbatim in \texttt{manual.md}, and on the rest \armC{}
returned prose the scorer cannot parse, or an ordinary near-miss.

\paragraph{On one model the manual is indistinguishable from a bare schema
on hard tasks.} \mSON{}'s \armM{} scores 79 of 332 hard tasks against
\armS{}'s 76: paired McNemar $p{=}0.72$, 17 discordant pairs one way and 14
the other. The same 24{,}177-character prompt that lifts \mGLM{} by 9 points
and \mDS{} by 20 lifts \mSON{} by less than one; the contract carrying the
same knowledge lifts it by 45.5. It is also the only model on which empty
governed tools beat the whole manual by a significant margin (38.0\%
against 23.8\%, hard split $p{=}6{\times}10^{-8}$, 15/62). One alternative is not excluded: that prompt
puts 191k input tokens per task in front of \mSON{} against \armS{}'s 91k,
and a context-load effect would produce the same null. The clause analysis
of Section~\ref{sec:clauses} cannot separate the two, since rules absent from
the SQL is what both predict.

\subsection{Content dominates scaffolding}
\label{sec:decomposition}

\begin{figure}[t]
\centering
\includegraphics[width=\columnwidth]{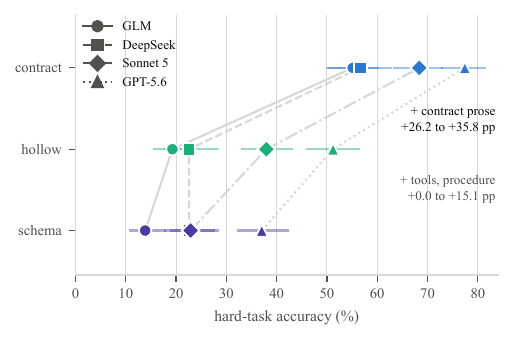}
\caption{The two steps of the decomposition on hard tasks, four models in
order of bare-schema capability. Adding the nine governed tools, the
retrieval instruction, the table allow-list and the operation rules with
the prose emptied (\armS{} to \armH{}) moves accuracy by $+5.4$, $+0.0$,
$+15.1$ and $+14.2$ points; restoring the prose (\armH{} to \armC{}) moves
it by a further $+35.8$, $+34.0$, $+30.4$ and $+26.2$. Bars are Wilson
95\% intervals.}
\label{fig:ladder}
\end{figure}

On hard tasks the gain splits into a scaffolding step, \armS{} to \armH{},
and a content step, \armH{} to \armC{}: 13.9 to 19.3 to 55.1 on \mGLM{},
22.6 to 22.6 to 56.6 on \mDS{}, 22.9 to 38.0 to 68.4 on \mSON{}, and 37.0 to
51.2 to 77.4 on \mGPT{} (Figure~\ref{fig:ladder}; the same rows in
Table~\ref{tab:headline}, and the last row of Table~\ref{tab:mcnemar}).

The content step is the larger of the two on every model. The scaffolding
step depends on the model: $+5.4$ on \mGLM{} ($p{=}0.058$ over all tasks,
20/35; on the hard slice alone $p{=}0.015$), $+0.0$ on \mDS{} ($p{=}0.61$,
19/15), $+15.1$ on \mSON{} ($p{=}9{\times}10^{-9}$, 14/64) and $+14.2$ on
\mGPT{} ($p{=}4{\times}10^{-7}$, 20/67). Three of the four models favour the
scaffolding and one is flat, and the two that favour it most are from
different model families.

The claim the hollow arm was built to test survives. The sceptical reading, that
the contract helps through a narrowed table surface, an imposed procedure,
or simply by making the agent slow down, predicts that \armH{} captures most
of the gain. It does not, on any model: the content step is $6.6{\times}$
the scaffolding step on \mGLM{}, the whole of the effect on \mDS{},
$2.0{\times}$ on \mSON{} and $1.8{\times}$ on \mGPT{}. \textbf{Scaffolding is a real
but secondary effect that the content dominates everywhere.} On the two
flash models alone this step would have read as a null;
Section~\ref{sec:power} says why that would have been weak evidence.

\paragraph{Empty tools are not free.} \armH{} is the most expensive arm on
both frontier models (\$46.08 on \mSON{}, \$22.09 on \mGPT{}), makes the
most tool calls on three of the four, burns the most reasoning on \mGLM{}
and \mSON{}, and forces answers on three models where \armC{} forces none
(Table~\ref{tab:efficiency}). The transcripts show why: the agent calls
\texttt{lookup\_domain}, is told nothing is available, and searches harder.
On \mSON{} it makes that call 392 times, plus 1{,}620 calls to
\texttt{lookup\_metric} with every metric emptied; on \mGPT{}, on 401 of
401 tasks. The
models that gain most from the scaffolding are also the ones that exercise
the lookup tools hardest.

\subsection{The compiled ceiling and the derivation gap}
\label{sec:ceiling}

Every result so far compares arms to each other. None says how much of the
contract's own content an agent recovers, because nothing establishes what
recovering all of it would look like. The contract lets us establish that.
Transcribing its \texttt{sql\_expression} fields into two DuckDB views (the
fee-rule match predicate, the natural-month reconstruction, the monthly
volume and fraud-level aggregates, the fee formula, and the sum over matched
transaction--rule pairs) gives a pre-computed fee layer of the kind a
context-layer product ships. Nothing in it is authored independently of the
contract; the only work the transcription adds is composition.

Which tasks the views cover is decided from question text alone, never
from gold and never from any arm's output. DABStep's tasks are instances of
parameterised question shapes (Section~\ref{sec:dabstep}), and five of
those shapes, in eleven date and grouping variants, ask exactly what the
views compute: the total fees a merchant paid in a period, the fee rules
that applied to it, the merchants a rule applies to, the average fee for a
transaction profile, and the rules matching an account type and ACI. A
regular expression per shape matches 176 of the 401 tasks, all hard.
Instantiating the views with each task's own parameters and grading the
result with the official scorer gives the ceiling: \textbf{the compiled
contract reproduces the benchmark's gold answer on every one of the 176
tasks it covers.} This is a statement about the artifact, not about an
agent, and it has two uses. It answers the objection that \armC{} wins
because the contract is subtly wrong in a benchmark-favouring way: it
compiles to gold. And it converts accuracy into recovery against a proven
ceiling. We call the 176 the \emph{macro} bucket. Of the remaining tasks,
148 need the same semantics plus a counterfactual or an optimisation the
views do not encode (\emph{derived}), and 77 involve no fee semantics. A
layer compiled from a domain contract's declared expressions therefore
reaches 53\% of this benchmark's hard set, a bound on that kind of layer
rather than on pre-computation in general (Section~\ref{sec:prior}). On
the 176, whatever an agent fails to get is a failure to derive
(Table~\ref{tab:buckets}; \appref{app:golds} says which of the 176 share a
premise with the gold verification).

\begin{table*}[t]
\centering
\small
\setlength{\tabcolsep}{4pt}
\begin{tabular}{@{}lrrrrrrrr@{}}
\toprule
& \multicolumn{2}{c}{\mGLM{}} & \multicolumn{2}{c}{\mDS{}} & \multicolumn{2}{c}{\mSON{}} & \multicolumn{2}{c}{\mGPT{}} \\
\cmidrule(lr){2-3}\cmidrule(lr){4-5}\cmidrule(lr){6-7}\cmidrule(lr){8-9}
Hard tasks & \emph{macro} & \emph{derived} & \emph{macro} & \emph{derived} & \emph{macro} & \emph{derived} & \emph{macro} & \emph{derived} \\
\midrule
\armS{} & 2.8\% & 25.0\% & 8.5\% & 34.7\% & 6.8\% & 37.8\% & 31.8\% & 41.2\% \\
\armH{} & 10.2\% & 25.7\% & 10.2\% & 31.7\% & 36.4\% & 37.2\% & 46.6\% & 54.1\% \\
\armM{} & 14.8\% & 32.4\% & 38.1\% & 44.6\% & 8.5\% & 38.5\% & 49.4\% & 48.6\% \\
\armC{} & \textbf{60.8\%} & \textbf{45.9\%} & \textbf{62.7\%} & \textbf{46.5\%} & \textbf{73.9\%} & \textbf{60.1\%} & \textbf{94.9\%} & \textbf{56.1\%} \\
\midrule
\emph{compiled contract} & \emph{100\%} & --- & \emph{100\%} & --- & \emph{100\%} & --- & \emph{100\%} & --- \\
\bottomrule
\end{tabular}
\caption{Hard tasks split by whether the compiled contract answers the task
outright (\emph{macro}, $n{=}176$; 118 on \mDS{}'s complete-case set) or
requires reasoning on top of it (\emph{derived}, $n{=}148$; 101). The gap
between \armC{} and the compiled contract on \emph{macro} is derivation
failure on information the agent provably had.}
\label{tab:buckets}
\end{table*}

Three readings follow.

\textbf{The contract's advantage concentrates where its semantics apply
directly.} Over \armS{} it is $+58.0$, $+54.2$, $+67.1$ and $+63.1$ points on
\emph{macro} against $+20.9$, $+11.8$, $+22.3$ and $+14.9$ on \emph{derived}.
This is the domain-specificity of Section~\ref{sec:domain}, measured on a
cleaner boundary.

\textbf{The two buckets do not move together.} \armC{} scores 60.8\%, 62.7\%,
73.9\% and 94.9\% on \emph{macro}, rising with capability (with one caveat:
\mDS{} and \mSON{} are 0.3 points apart on the capability axis and 11 apart
here), against 45.9\%, 46.5\%, 60.1\% and 56.1\% on \emph{derived}, which
rises under no ordering: \mSON{} leads it while trailing \mGPT{} by 21
points on \emph{macro}.

\textbf{The derivation gap closes with capability.} Read against the
ceiling, the \emph{macro} series is a shortfall of 39.2, 37.3, 26.1 and 5.1
points below an accuracy proven reachable: the \emph{derivation gap}, the
price of making an agent derive a query the contract could have
pre-computed. Against a ceiling of 100\% the gap is arithmetically the arm's
error rate, and ``errors fall as models improve'' is not a finding. The
finding is the contrast with \emph{derived}, whose shortfall we call the
\emph{composition gap}: between the flash models and the strongest one, the same capability increment closes 87\% of the
macro-bucket gap and 19\% of the derived-bucket one, and \mSON{}'s
\emph{macro} score falls between the flash models and \mGPT{} while its
\emph{derived} score exceeds all three. We recorded the direction of this
result publicly before \mGPT{} ran, predicting a \emph{macro} score ``well
above 62.7\%, moving toward 100\%'' if capability governed it and ``near
61--63\%'' if the harness did; the observed value is 94.9\%. \mSON{}, run
last and outside the prediction, lands at 73.9\%: the direction replicated,
the size did not.

This bears on the economics of Section~\ref{sec:intro}. The derivation gap
is exactly what separates a declarative contract, which requires the agent
to derive the query, from pre-computed macros, which do not. A gap fixed
near 38 points would be a standing structural cost of the declarative
approach. A gap that falls to 5 points on the strongest model is a
\emph{transient} cost, one the trend in model capability is already paying
down, while the per-metric authoring cost of a macro layer is paid down by
nothing. What survives the strongest model is not difficulty in knowing the
domain but the counterfactual and optimisation reasoning built on top of it,
on which the four models have no clean ordering.

\subsection{The gain stays inside the contract's domain}
\label{sec:domain}

DABStep's hard set is dominated by one question family: 294 of 332 hard
tasks concern fees, and the contract encodes fee semantics. On those 294 the
contract arm scores 59.2\%, 61.2\%, 73.5\% and 83.7\% against the manual
arm's 24.5\%, 44.9\%, 24.1\% and 53.4\%. On the 38, 30, 38 and 38 hard tasks
that do not concern fees it holds no detectable advantage (9 against 4, 8
against 9, 11 against 8 and 11 against 10 correct), and on \mGPT{} the bare
schema leads the slice outright at 14, the only slice in the four runs where
it does (\appref{app:domain}). Thirty of the 38 are one family, which needs
the fee rules without naming them and scores 0 to 3 of 30 for every arm on
three of the four models. The slices are too small to be a negative result
and equally unable to support a general claim. ``Hard accuracy'' on this
benchmark is close to ``fee-question accuracy'', so the result should be
stated as: \textbf{a contract carrying a domain's semantics produces a large
gain on questions in that domain.} That is what a domain contract should do,
and it is why this benchmark cannot answer the general question.

\section{Why the Contract Arm Wins}
\label{sec:why}

Aggregate accuracy does not explain itself. This section traces one task end
to end, reads the SQL every arm submitted, accounts for the work and money
each arm spent, and looks at where the contract arm still loses.

\subsection{The rule was in both prompts; only one arm's SQL carries it}
\label{sec:reach}
\label{sec:clauses}

Task~1278 asks for \emph{the average fee NexPay would charge for a credit
transaction of 50 EUR}.

\begin{center}
\footnotesize
\setlength{\tabcolsep}{4pt}
\begin{tabular}{@{}lr@{}}
\toprule
Gold      & \texttt{0.352294} \\
\armC{} (correct)  & \texttt{0.352294} \\
\armM{} (wrong)    & \texttt{0.353053} \\
\armS{} (wrong)    & \texttt{0.353053} \\
\midrule
\texttt{is\_credit IS NULL OR is\_credit = true} & \texttt{0.352294} \\
\texttt{is\_credit = true} & \texttt{0.353053} \\
\bottomrule
\end{tabular}
\end{center}

Both figures reproduce against the warehouse with the two predicates
shown. A \texttt{NULL} in a \texttt{fees} column means \emph{this rule
applies to every value}, not \emph{unknown}; both baselines dropped the
\texttt{NULL} rows and produced the identical wrong number. The contract arm called
\texttt{lookup\_domain("fees")}, and its recorded reasoning states the rule
back: \emph{``rules that apply to credit transactions: is\_credit true or
null''}.

The decisive detail is that \texttt{manual.md} documents this rule too.
\armM{} had it, inside a 24{,}177-character system prompt. \armC{}'s system
prompt is 2{,}475 characters, with the rule reachable by one targeted call.
Two published results frame the difference: retrieval quality falls for
material buried mid-prompt~\cite{lost-in-the-middle}, and what a tool's
documentation says changes what an agent does with
it~\cite{tool-documentation}. \armM{}'s copy of the rule was buried
mid-prompt; \armC{}'s was a tool's documentation.

The finding is not that the contract arm was given more information: for
this rule both arms had it, and only one used it. What the contract changes
is where the rule sits at the moment of use, in the result of one targeted
call rather than in a long prompt. The scoping matters: it holds for the
clauses the manual states outright, which is the class this failure
belongs to, and not for the three clauses the contract resolves on its own
(Section~\ref{sec:interpretation}).

One task shows the mechanism. Whether it is general can be read from the
SQL every arm submitted, rather than from the answers. We defined one detector per clause of the contract's fee semantics
that a correct query must express: the \texttt{NULL}-wildcard disjunct, the
empty-list wildcard, the capture-delay band mapping, a per-merchant monthly
aggregate, the natural-month reconstruction, and fraud measured on euro
volume. We then restricted the comparison to the 97 tasks that require all
six (69 on \mDS{}, whose truncation cost the rest), so every row compared
carries the same requirement (Table~\ref{tab:clauses}). The detectors are
deliberately permissive: the question is whether the agent expressed the
idea, not whether it copied the contract's phrasing.

\begin{table}[t]
\centering
\footnotesize
\setlength{\tabcolsep}{3pt}
\begin{tabular}{@{}lrrrr@{}}
\toprule
\tabhead{Wrote all six clauses} & \mGLM{} & \mDS{} & \mSON{} & \mGPT{} \\
\midrule
\armS{} & 7\% & 4\% & 3\% & 8\% \\
\armH{} & 14\% & 4\% & 4\% & 8\% \\
\armM{} & 3\% & 9\% & 4\% & 4\% \\
\armC{} & \textbf{39\%} & \textbf{65\%} & \textbf{55\%} & \textbf{98\%} \\
\bottomrule
\end{tabular}
\caption{Share of the 97 tasks requiring all six fee clauses (69 on \mDS{})
on which the agent's submitted SQL expresses all six. \armM{} holds the same
knowledge as prose in its system prompt. \armC{} against \armM{} is
$p{=}2{\times}10^{-10}$, $3{\times}10^{-12}$, $9{\times}10^{-16}$ and
$2{\times}10^{-47}$ (Fisher exact).}
\label{tab:clauses}
\end{table}

\textbf{Note what does not move.} \armM{} writes no more of the contract's
clauses on a strong model than on a weak one (3\%, 9\%, 4\%, 4\%), while
\armC{} reaches 55\% and 98\% on the two frontier models. \mSON{} is the
extreme case: 4\% against 55\% in the SQL, no better than a bare schema
against 45 points above it in accuracy. Capability does not, on its own,
make a model extract structure from prose; it makes a model much better at
\emph{using} structure it is handed. That is the same asymmetry the
derivation gap shows (Section~\ref{sec:ceiling}), measured on behaviour
instead of outcomes.

\textbf{What this does not measure.} Clause presence does not predict
success: within every arm on every model, attempts that got the task right
and attempts that got it wrong write the same clauses (every Fisher
$p{\ge}0.14$), and \armH{} on \mGLM{} writes more clauses than \armM{} while
scoring at the bare-schema floor. The measure captures whether the
contract's vocabulary reaches the query, not whether the query is any good.

\subsection{Less work, and usually less money}
\label{sec:efficiency}

\begin{table*}[t]
\centering
\small
\begin{tabular}{@{}lrrrr@{}}
\toprule
 & \armS{} & \armH{} & \armM{} & \armC{} \\
\midrule
\multicolumn{5}{@{}l}{\emph{\mGLM{}}}\\
Turns / task & 12.1 & 8.4 & 9.8 & \textbf{7.1} \\
Tool calls & 5{,}134 & 4{,}110 & 4{,}282 & \textbf{3{,}355} \\
Reasoning tokens & 428k & 525k & 400k & \textbf{154k} \\
Cost / correct & \$0.0084 & \$0.0080 & \$0.0076 & \textbf{\$0.0029} \\
Forced answers & 41 & 0 & 18 & \textbf{0} \\
\midrule
\multicolumn{5}{@{}l}{\emph{\mDS{}}}\\
Turns / task & 13.7 & 11.4 & 12.8 & \textbf{8.2} \\
Tool calls & 5{,}489 & 6{,}375 & 4{,}770 & \textbf{4{,}155} \\
Reasoning / task & 10{,}941 & 10{,}766 & 9{,}243 & \textbf{4{,}844} \\
Cost / correct & \$0.0195 & \$0.0173 & \$0.0117 & \textbf{\$0.0061} \\
Forced answers & 49 & 15 & 38 & \textbf{0} \\
\midrule
\multicolumn{5}{@{}l}{\emph{\mSON{}}}\\
Turns / task & 11.3 & 10.0 & 11.1 & \textbf{7.0} \\
Tool calls & 5{,}644 & 6{,}494 & 5{,}176 & \textbf{3{,}791} \\
Reasoning tokens & 942k & 1{,}411k & 1{,}044k & \textbf{824k} \\
Cost / correct & \$0.288 & \$0.259 & \$0.311 & \textbf{\$0.136} \\
Forced answers & 43 & 27 & 32 & \textbf{0} \\
\midrule
\multicolumn{5}{@{}l}{\emph{\mGPT{}}}\\
Turns / task & 7.9 & 10.4 & 7.6 & \textbf{6.5} \\
Tool calls & 5{,}099 & 7{,}997 & \textbf{4{,}354} & 5{,}245 \\
Reasoning tokens & 357k & 311k & 277k & \textbf{141k} \\
Cost / correct & \$0.082 & \$0.099 & \textbf{\$0.062} & \$0.067 \\
Forced answers & 0 & 8 & 0 & \textbf{0} \\
\bottomrule
\end{tabular}
\caption{Work and cost per arm. The contract arm uses the fewest turns and
the least reasoning on every model and never exhausts its output budget; it
is the cheapest arm per correct answer on three of the four. \mDS{}'s
reasoning row is per task rather than a total, because its complete-case set
has fewer rows than the other runs.}
\label{tab:efficiency}
\end{table*}

\armC{} is not thinking harder; it has less to search for. On all four
models it uses the fewest turns and the least reasoning (by a factor of two
to three on three models, by 12\% on \mSON{}), and it is the only arm that
never exhausted its output budget on any model (Table~\ref{tab:efficiency};
on reporting cost beside accuracy, see~\cite{agents-that-matter}).

Money follows work on three models and not on the fourth. \armC{} is the
cheapest arm outright on both flash models, and per correct answer it is the
cheapest on \mGLM{}, \mDS{} and \mSON{}, by about $2{\times}$ on \mSON{}
(\$0.136 against \$0.259 for the next arm). On \mGPT{} it bills \$21.32
against \armM{}'s \$14.42, and \$0.067 per correct answer against \$0.062.
The recorded token counts locate the difference (Table~\ref{tab:gptcost}):
fresh, uncached input accounts for \$5.3 of the \$6.9 gap and output for
\$1.6, and cached input is a wash. \armC{} reads 1{,}692 fresh tokens per
turn against \armM{}'s 582, because its knowledge arrives as tool results.
On \mGPT{} it calls \texttt{lookup\_metric} 1{,}690 times and
\texttt{lookup\_domain} 776 times over 401 tasks, two tool calls per turn,
and each result is billed at the fresh rate when it first enters the
context. \armM{}'s knowledge sits in a system prompt that is the same prefix
in every request of every task, and is billed at the cached rate. \armC{}
also emits more visible output (726k tokens against 425k) despite half the
reasoning (141k against 277k), because \mGPT{} validates before it runs: 676
\texttt{inspect\_query} calls against 622 \texttt{run\_query}, so most
queries are written twice. \mSON{} shows the same mechanism (1{,}976 against
1{,}037 fresh tokens per turn), but its \armM{} arm carries 191k tokens of
context per task and pays \$14.40 in cached input alone, which is what makes
the contract arm cheaper there.

\begin{table}[t]
\centering
\footnotesize
\setlength{\tabcolsep}{5pt}
\begin{tabular}{@{}lrrrr@{}}
\toprule
\mGPT{} & \tabhead{Fresh input} & \tabhead{Cached input} & \tabhead{Output} & \tabhead{Total} \\
\midrule
\armM{} & \$3.55 & \$3.85 & \$7.02 & \$14.42 \\
\armC{} & \$8.87 & \$3.78 & \$8.67 & \$21.32 \\
\bottomrule
\end{tabular}
\caption{\mGPT{}'s bill for the two content-bearing arms, decomposed from the
recorded token counts at the pinned endpoint's prices: \$2.00 per million
fresh input tokens, \$0.20 cached, \$10.00 output (reasoning included).}
\label{tab:gptcost}
\end{table}

The general statement is about prices, not about the contract: at a
ten-to-one ratio between fresh and cached input, on-demand delivery is
charged per task and a static prompt once per cache window, and whether the
contract's savings in turns and reasoning cover the difference depends on
how often the model looks things up and how long its conversations run.
What generalises across the four models is the work claim, not the price
claim.

\subsection{Where the contract arm fails}
\label{sec:losses}

On \mGLM{} \armC{} loses 14 tasks to \armS{} and 29 to \armM{}, 10 to both,
and two modes account for nearly all of them: set membership under
\texttt{NULL} wildcards, the rule of Section~\ref{sec:reach} failing in the
other direction (on task~1500, \emph{fee IDs that apply to account\_type = O
and aci = C}, \armC{} returns both false positives and false negatives while
\armM{} matches exactly), and near-miss arithmetic (task~1274: gold
\texttt{0.126459}, \armC{} \texttt{0.125516}). These are not retrieval
failures. \texttt{lookup\_domain} is called on 331 of 401 tasks, the 70 that
skip it are easier for every arm, and 26 of the 33 losses occur on tasks
where the lookup was performed; \mSON{} repeats the pattern: of the 42 tasks it
loses to at least one other arm, 35 had the lookup. The agent fetches the rule and misapplies it.
That points away from hoisting rule bodies into the system prompt, since
the rules were already fetched, and toward making the wildcard semantics
executable: a
five-line DuckDB macro encoding the rule the contract states in prose
answers tasks~1500 and~1502 exactly. We did not build that tool before
measuring: it would have been designed from knowledge of which tasks the
contract arm lost, and an artifact frozen before any question was read is
this study's strongest methodological asset.

\section{Governance}
\label{sec:governance}

The contract declares forbidden operations as well as semantics, and the
tool layer enforces them. Because the same runs record every statement the
agent submitted and a post-run integrity check on the database, a second
result falls out of the same 6{,}416 transcripts. We counted mutating
statements the model \emph{submitted} to a SQL tool: \texttt{CREATE
TABLE}/\texttt{VIEW}, \texttt{DROP}, \texttt{INSERT}, \texttt{UPDATE},
\texttt{DELETE}, \texttt{ALTER} or \texttt{TRUNCATE} at statement position in
the query argument (Table~\ref{tab:governance}).

\begin{table}[t]
\centering
\small
\setlength{\tabcolsep}{4pt}
\begin{tabular}{@{}lrrrr@{}}
\toprule
 & \armS{} & \armH{} & \armM{} & \armC{} \\
\midrule
\mGLM{} & 27 (8) & \textbf{0} & 92 (13) & \textbf{0} \\
\mDS{} & 24 (2) & \textbf{0} & 1 (1) & \textbf{0} \\
\mSON{} & \textbf{0} & \textbf{0} & 22 (2) & \textbf{0} \\
\mGPT{} & \textbf{0} & \textbf{0} & \textbf{0} & \textbf{0} \\
\midrule
\textbf{Total} & \textbf{51 (10)} & \textbf{0} & \textbf{115 (16)} & \textbf{0} \\
\bottomrule
\end{tabular}
\caption{Mutating statements submitted, by arm (tasks in parentheses).
Fisher exact on traces
containing at least one: $p{=}8{\times}10^{-7}$ for \mGLM{} alone,
$p{=}3{\times}10^{-8}$ pooled over the three models where any arm attempted a
write. \mDS{} and \mSON{} alone are underpowered ($p{=}0.25$ and $0.50$).}
\label{tab:governance}
\end{table}

\textbf{Ungoverned arms submitted 166 mutating statements across 26 tasks;
governed arms submitted none, on all three models where any arm attempted a
write.} \mGPT{} contributes nothing in either direction: handed a raw SQL
tool and no rules, it never attempted a write, so there was nothing to
deter. \mSON{} supplies the hazard \mGPT{} lacks, with 22 statements on two
\armM{} tasks, one of which reached the database, so the contrast has a
frontier model on both sides. The idiom rather than the capability predicts
the attempt: \mGPT{} holds intermediate results in common table expressions
on 59\% and 55\% of its ungoverned traces against 26--37\% for the other
three models, and never issues \texttt{CREATE TEMP}, the statement behind every
mutation attempt on \mGLM{}.

The governed arms have two defences, both declared as data in the contract:
the forbidden operations and the table allow-list are rendered into the
system prompt, and the same two fields drive the static checkers that
validate every statement before execution (we confirmed against a scratch
copy that each mutating statement type is refused with an explicit
violation while an ordinary \texttt{SELECT} executes). \textbf{The
experiment exercised only the first.} A rejected tool call still appears in
the transcript, so the count above is of what the model emitted, and zero
means \emph{never attempted}, not attempted and blocked. In 6{,}416 rows the
enforcement layer never had to fire against a mutating statement in a
governed arm, because the declared rules stopped the attempts upstream of
it. That is the declarative mechanism RuLES found unreliable~\cite{rules},
measured here at 166 to 0; it costs nothing at query time and depends on
compliance, and the data cannot say whether a model that would not comply
would have been caught, because none tried.

Two escalations reached the database: task~68 (\armS{} on \mDS{}) and
task~2546 (\armM{} on \mSON{}). \mDS{} created a \texttt{TEMP} table, lost it on the next call because the
ungoverned tool opens a fresh connection per call, diagnosed that in its
reasoning, and switched to persistent tables; \mSON{} made the same
diagnosis on one task and, on another, went straight to persistent tables.
Both answered the question wrongly as well (\appref{app:escalations} quotes
the transcripts). Neither is an
injection or a jailbreak. Each is a competent agent routing around a session
boundary to finish its work, in which every statement is locally reasonable
and no test written in advance would have anticipated the path, which is the
argument for a declarative rule that does not need to. Two limits follow.
The size of the hazard is partly a property of this harness, since a
session-scoped connection would likely have prevented both events; the
contrast survives because the governed arms faced the identical harness and
emitted nothing. And two corruptions are an anecdote, not a rate: the
defensible number is the 166 attempts, and the defensible claim is that on
models which reach for the mutating idiom, declared rules stopped them from
trying.

\section{Threats to Validity}
\label{sec:threats}

\subsection{One benchmark, one domain}

The gain is confined to fee questions, 294 of the 332 hard tasks
(Section~\ref{sec:domain}), and the benchmark is a concentrated one: 450
tasks reduce to 113 question shapes, 27 of which cover three quarters of it
(Section~\ref{sec:dabstep}). Results on so narrow a question space should
not be read as results about analytics agents in general. We ran no second
benchmark; BIRD's withheld evidence~\cite{bird} is the nearest analogue,
but it supplies a hint per question rather than a domain manual.

\subsection{The contract resolves ambiguities the manual leaves open}
\label{sec:interpretation}

The arms are matched on the sources they may consult, not on how those
sources read. Seven metric descriptions carry an \texttt{INTERPRETATION}
sentence (Section~\ref{sec:contract}). Four restate or apply what the manual
supplies; three go further. The manual says a \texttt{null} field ``applies
to all possible values'', but the annexed \texttt{fees} data encodes that
wildcard for list-typed fields as an \emph{empty list}, which the manual
never mentions, and two band metrics break a tie on a shared boundary that
the manual's ``between A and B'' does not decide. All three are recoverable
from data every arm could query, and none supplies a task's answer; but they
were recovered once, by us, and handed to one arm, and the empty-list clause
matters: reversing it in the compiled contract changes the answer on 87 of
the 97 tasks in the payments families. The contract arm's margin therefore
combines delivery of what the manual states with resolution of what it
leaves open. Separating them needs a fifth arm, the contract with its seven
clauses stripped, which we did not run.

\subsection{No repeats, and two frontier points that disagree}
\label{sec:power}

None of the four runs repeats. One near-replicate exists: the all-450
\armC{} sweep on \mGLM{} submitted to the leaderboard
(Section~\ref{sec:golds}) shares 401 tasks with the \mGLM{} run and differs
from it only by a validator fix. Paired, the two agree on accuracy (60.3\%
against 57.9\%, exact McNemar $p{=}0.35$) and \textbf{disagree on 94 of 401
tasks, a 23.4\% flip rate}, at temperature 0 on a pinned endpoint. The
churn is symmetric (52 one way, 42 the other), so the McNemar contrasts,
which read asymmetry, are not at risk; but no claim about a named task's
verdict should be read as reproducible, and the frontier runs, which could
not pin temperature, are likelier to flip more. Table~\ref{tab:mcnemar}
reports 24 tests and corrects none; the three vs-\armC{} contrasts survive
any correction, and the scaffolding row on the two flash models does not.
Underpowered comparisons are the norm in this
literature~\cite{little-power}.

The capability claims rest on two frontier models that disagree, by 21
points on the \emph{macro} bucket and by 4, the other way, on
\emph{derived}, and the literature does not settle the direction
either~\cite{prompt-engineering-advanced,scaffold-gaia}. Capability is also
not one number here: \mDS{} and \mSON{} tie on the bare-schema axis and sit
12 points apart with the contract. With only the two flash models, the
scaffolding step was a null (pooled $p{=}0.29$), the prompt margin shrank
with capability, and the derivation gap sat fixed near 38 points; \mGPT{},
the third model run, overturned all three, putting the scaffolding step at
$+14$ points at $p{<}10^{-6}$, and \mSON{}, run last, confirmed it at
$+15$. \textbf{A null across two adjacent models is weak evidence of
absence}, and the size of the derivation gap, which rests on two frontier
points 21 apart, deserves the same discount.

\subsection{Reconstructed golds are lenient}
\label{sec:threats-golds}

Every accuracy here except the officially graded pair rests on reconstructed
golds that are one-sidedly lenient on the hard split
(Section~\ref{sec:golds}); read every absolute hard-split figure as a few
points optimistic. The proxy that bounds the other cells moves the twelve
\armC{}-versus-baseline gaps by $-2.8$ to $+1.4$ points against raw gaps of
13.7 to 45.5 (\appref{app:leniency}). No conclusion here turns on 3 points.

\subsection{Contamination}

DABStep is public, and we ran no contamination probe. A memorising model
would show what we report: high accuracy on the \emph{macro} bucket, whose
tasks follow the benchmark's most common shapes, and less on \emph{derived}.
Against that, \mGPT{}'s contract arm scores 94.9\% on the same tasks where
its bare-schema arm scores 31.8\%, and \mSON{}'s 73.9\% against 6.8\%, and
memorisation should not care which context the agent was handed. That is an
inference, not a measurement.

\section{Conclusion}
\label{sec:conclusion}

Context layers work; the open question was which part of one does the
work, and it matters because the parts cost different amounts. Holding the
tool surface, the retrieval instruction, the table allow-list and the
operation rules byte-for-byte fixed and emptying only the prose,
\textbf{the content step dominates the scaffolding step on every model}, by
$1.8{\times}$ to $6.6{\times}$. The contract beats the same knowledge
pasted into a prompt because its rule is one lookup away at the moment of
use, visibly so in the SQL it submits, and it does so with the fewest turns
and the least reasoning on every model and at the lowest cost per correct
answer on three of four. Because the contract compiles to a layer that
reproduces gold, each arm's accuracy is a recovery rate: the contract arm
recovers 60.8\%, 62.7\%, 73.9\% and 94.9\% of what it was handed, in order
of capability, so \textbf{the cost of making an agent derive rather than
call a macro is transient}, while the composition gap above it does not yet
track capability. Declared rules the model merely reads stopped every
mutating statement, 166 to zero, without the enforcement layer behind them
ever firing.

Three limits bound all of this: the gain appears only inside the domain the
contract describes, the benchmark is a single and concentrated one, and no
run repeats, so the aggregate comparisons stand and no individual verdict
is reproducible.

\paragraph{Future work,} in order of what it would settle. Repeat runs on
the two models whose temperature can be pinned would put a variance
estimate under every number here. A fifth arm, the contract with its seven
interpretation clauses stripped, would separate delivery of what the manual
states from resolution of what it leaves open. And an executable metric
layer, a tool that runs a declared expression rather than describing it,
would separate the declarative contribution from the executable one the way
this paper separates content from scaffolding; MotherDuck's later
with/without comparison on that component, a generated semantic layer at
95\% against 100\% for Markdown plus SQL at $2.5{\times}$ the
tokens~\cite{motherduck-ai-writes}, lands where our ablation does. The
order is measure, ship, re-measure; this paper is the baseline.

\section*{Artifact Availability}

The harness, the frozen contract and its hollow variant, and the raw result
rows are tracked at
\url{https://github.com/flyersworder/agentic-data-contracts}; all 6{,}416
transcripts including model reasoning are attached to the release tagged
\texttt{dabstep-eval-2026-09} there, with SHA-256 digests. Every row records the commit, contract
digest, gold-set hash, scorer, endpoint, token counts, cost and a post-run
integrity check.

% Content ends here; `make check` reads this label's page from the .aux
% to count content pages against the PVLDB limit.
\label{end:content}

\bibliographystyle{abbrv}
\bibliography{refs}

\ifextended
\appendix
\section{Gold Reconstruction and External Validation}
\label{app:golds}

DABStep withholds golds for its 450 test tasks, which is good for the
benchmark and inconvenient for anyone running an offline ablation. We
reconstruct them from the leaderboard's own published artifacts, and because
this is the one component a reader cannot check independently, we describe
it in full.

Every submission to the DABStep leaderboard publishes a per-task file
containing, for each task, the submitted answer and whether the official
grader marked it correct. At the time of reconstruction the corpus held
2{,}199 such submissions. Reconstruction proceeds by plurality over
\emph{officially-graded-correct} answers only:

\begin{enumerate}
\itemsep2pt
\item An answer enters the vote only if DABStep's own grader marked that
      submission correct on that task. Crowd agreement alone is never
      sufficient; the benchmark's own verdict is the admission criterion.
\item Answers are normalised with the benchmark's own normaliser, so
      submissions differing only in formatting or list order agree.
\item A task yields a gold only if one answer holds at least a 75\%
      plurality of admitted votes. Median realised share: 0.981.
\end{enumerate}

After excluding five tasks whose consensus answer manual verification showed
to be wrong, this yields golds for 401 of 450 tasks (332 hard, 69 easy).
Consensus inherits the leaderboard's errors, and we found some.

\paragraph{Independent verification.} Consensus is an argument, not a
measurement, so we checked it against the database. The largest template
families, the three fee-average question shapes, were recomputed directly
from DuckDB using the matching rule documented in \texttt{manual.md}. All 59
reconstructed golds reproduce exactly.

\paragraph{What the compiled ceiling shares with this check.} Of the 176
tasks the compiled contract answers (Section~\ref{sec:ceiling}), the 97 in
the seven payments families exercise the full
\texttt{payments}-to-\texttt{fees} join the views encode; the other 79
resolve off \texttt{fees} and \texttt{merchant\_category\_codes} alone, with
the same wildcard predicates but not the join. And 59 of those 79 are
exactly the tasks verified above using the matching rule the contract
declares, so on that subset the ceiling and the gold check share a premise;
the other 117, graded against consensus golds we did not recompute, carry
the claim.

\paragraph{External validation.} A leaderboard submission converts the
argument into a measurement, because every submission returns per-task
official verdicts on the exact answers we scored against reconstructed
golds. We ran that check on both arms of the main contrast. Sweeps of
\armC{} and \armM{} over all 450 tasks on \mGLM{}, at the same commit, the
same pinned provider endpoint and the same harness, were submitted and
graded against the withheld golds. (The \armM{} sweep was interrupted and
resumed; three hard tasks it stamped \texttt{db\_corrupted} were re-run
serially and merged, changing no verdict.)

\begin{center}
\small
\begin{tabular}{@{}lrrrr@{}}
\toprule
split & $n$ & \armC{} & \armM{} & gap \\
\midrule
hard & 378 & \textbf{51.9\%} & \textbf{18.8\%} & $+33.1$~pp \\
easy & 72  & 70.8\% & 75.0\% & $-4.2$~pp \\
all  & 450 & 54.9\% & 27.8\% & $+27.1$~pp \\
\bottomrule
\end{tabular}
\end{center}

\noindent Paired McNemar on the hard split gives 137 tasks \armC{} alone
answers correctly against 12 for \armM{}, exact two-sided
$p{=}4.9\times10^{-28}$. The easy-split reversal, the effect living entirely
in the hard tasks, is confirmed externally too. \armM{}'s 18.8\% sits just
below the cluster of named manual-in-prompt baselines, 1.0~pp under its
lowest member, Claude~4 Sonnet at 19.8\%, and is the same reimplementation
graded the way those entries were rather than by us. The second submission
went up as an explicitly named baseline control after validation had closed,
so no ranking claim was available to buy.

The excluded tasks are not a biased sample. Officially, the 332 hard tasks
we keep score 52.4\% and the 46 hard tasks we drop score 47.8\%: the
consensus threshold filters on gold agreement, not on task difficulty, and
the 401-task subset stands in fairly for the 450.

The reconstructed golds are \emph{lenient}, one-sidedly, and only on hard
tasks. Across the 401 shared tasks the two gradings agree on 382 (95.3\%).
All 19 disagreements run the same way, correct under our grading and wrong
under the official one, and none run the other way. Easy is exact (49/69
under both); hard is 193/332 against 174/332 official, $+5.7$~pp. The same
measurement on \armM{} gives $+3.0$~pp, likewise one-sided.

The 19 look like formatting and are not. Our scorer is the leaderboard's own
file, vendored verbatim and hash-pinned, so both gradings apply identical
rules; re-running it on each disagreement against the reconstructed gold
returns \emph{correct} on all 19, where the official grading returns wrong
on all 19. The rules cannot be the difference, and indeed cannot fail on
these shapes: numeric comparison rounds both sides to the smaller precision,
so a full-precision answer against a rounded gold always matches, and
single-word comparison matches \texttt{D:219.36} against a bare \texttt{D}.
The 19 are wrong \emph{values} that the reconstructed golds concealed, not
wrong renderings of right ones. \appref{app:leniency} bounds what this
does to the results.

\section{The \mDS{} Run Lost 29\% of Its Rows}
\label{app:truncation}

Roughly 1{,}000 rows into the \mDS{} run, the pinned provider endpoint
began returning HTTP~429. The harness had no retry for that status, so each
throttled request was written to the results file as a terminal error: 466
of 468 error rows are rate limits. Error rate was 0.2\% over the first
1{,}000 rows, 31\% over the next 200, and 100\% thereafter. The sweep then
finished early, because a throttled request fails instantly, and exited
cleanly.

The loss costs power, not validity, and the evidence for that is specific:

\begin{itemize}
\itemsep2pt
\item Errors are near-uniform across arms: 113, 116, 118 and 121 of 401
      rows. No arm is preferentially damaged.
\item 114 of the 122 lost tasks lost \emph{all four} arms together, because
      the runner schedules a task's arms close in time. Independent per-arm
      failure would have left roughly 100 complete tasks; 279 survived.
      Missingness is a property of \emph{when} a task ran.
\item Figures on the 279 tasks scoreable in all four arms (the
      complete-case set, used for \mDS{} throughout) and per-arm figures
      over each arm's own denominator agree to within one point.
\end{itemize}

We therefore report \mDS{} on the 279 tasks scoreable in all four arms,
with intervals roughly 20\% wider than the other runs'. One residual caveat:
the lost tasks are slightly enriched for hard tasks (87\% hard against 83\%
overall), so \mDS{}'s absolute accuracies are marginally optimistic. Every
arm is affected identically. The 122 tasks are re-runnable, but only on a
different provider than the other 279, which would be its own confound;
leaving them out is cleaner.

One lesson generalises beyond this paper. Pinning a single provider endpoint
with fallbacks disabled is correct for validity, since it holds
quantization, price and throughput fixed across arms, but it transfers
responsibility for transient failures to the harness, and here nothing
looked wrong until the results were read. Benchmark harnesses that pin
endpoints need a rate-of-progress check, not just a liveness check. The
harness now retries 429.

\section{Scorer Leniency and the Correction}
\label{app:leniency}

Every accuracy in the paper except the officially graded pair of
\appref{app:golds} is measured against reconstructed golds, which the
leaderboard submissions show to be one-sidedly permissive: 19 answers graded
correct here are wrong officially and none go the other way, $+5.7$~pp on
the hard split for \armC{} and $+3.0$~pp for \armM{}. What remains exposed is
every absolute number for the other two arms and the other three models, and
the 20--26\% leaderboard band where it is quoted against a self-graded
figure.

The sharper question is whether the leniency is arm-neutral. If it is, it
moves levels and not contrasts. That is checkable from a proxy. Every one of
the 19 failed exact string match against the reconstructed gold and was
rescued by the scorer's tolerance, so exact-match failure among
locally-correct rows upper-bounds an arm's exposure
(\texttt{analysis/leniency.py}; exact match strips whitespace and Markdown
bold and nothing else, deliberately not the benchmark normaliser, which is
the tolerance being measured). On the hard split, where all 19 disagreements
lie, the flag has perfect recall at 33 flags, so a flag converts to a real
error at $19/33 = 0.576$.

\begin{table}[htbp]
\centering
\caption{Exposure to scorer leniency: the share of each arm's correct
hard-split answers that required more than exact match against the
reconstructed gold (exp.), and the accuracy that follows from correcting it
(corr.). An upper bound.}
\label{tab:leniency}
\footnotesize
\setlength{\tabcolsep}{3pt}
\begin{tabular}{@{}lrrrrrrrr@{}}
\toprule
& \multicolumn{2}{c}{\mGLM{}} & \multicolumn{2}{c}{\mDS{}}
& \multicolumn{2}{c}{\mSON{}} & \multicolumn{2}{c}{\mGPT{}} \\
\cmidrule(lr){2-3}\cmidrule(lr){4-5}\cmidrule(lr){6-7}\cmidrule(lr){8-9}
\tabhead{Arm} & exp. & corr. & exp. & corr. & exp. & corr. & exp. & corr. \\
\midrule
\armC{} & \textbf{15.8} & 50.1 & \textbf{17.2} & 51.0 & \textbf{10.6} & 64.2 & \textbf{10.5} & 72.7 \\
\armH{} & 26.6 & 16.3 & 21.6 & 19.8 & 14.3 & 34.8 & 14.7 & 46.9 \\
\armM{} & 19.7 & 20.3 & 23.7 & 37.1 & 40.5 & 18.2 & 13.8 & 46.3 \\
\armS{} & 37.0 & 10.9 & 31.4 & 18.5 & 36.8 & 18.0 & 10.6 & 34.8 \\
\bottomrule
\end{tabular}
\end{table}

\armC{} is the \emph{least} exposed arm on all four models, which is the
same mechanism the rest of the paper reports, surfacing in the grading
rather than the SQL: the contract states an answer format and the arms
without it improvise one. It does not follow that the correction is
conservative for the main contrast. \armC{} has far more correct answers to
lose, so on \mGLM{} and \mGPT{} it gives up the most \emph{points} (5.0 and
4.7), and on the other two \armM{} does. Across the twelve \armC{}-versus-baseline
contrasts the gap narrows on nine and widens on three, by $-2.8$ to $+1.4$~pp
against raw gaps of 13.7--45.5~pp. It widens exactly where the baseline is
the heavily exposed one: \mSON{}'s \armM{} sits at 40.5\% exposure, and
correcting both sides moves that contrast from 44.6 to 46.0~pp. That 40.5\%
also makes \mSON{}'s corrected \armM{} figure of 18.2\% the least safe
extrapolation in the analysis, applying a conversion rate measured on
\armC{} and \mGLM{} to the arm and model furthest from it; read it as a
direction, not a second measurement. The honest summary is not that the
leniency favours either side, but that its effect on every contrast is under
3~pp and no conclusion in the paper turns on 3~pp.

The correction's weakness is that it applies a conversion rate calibrated
on one arm to the others. The second submission tests exactly that.
Calibrating on \armC{} as before and applying the resulting rate blind to
\armM{} predicts a $3.1$~pp reconstructed-gold bias against a measured
$3.0$~pp. The point estimates agree, but on 18 flags the realised conversion
($10/18$) carries a Wilson 95\% interval of roughly $[0.34, 0.75]$, so any
measured bias between about $1.8$ and $4.1$~pp would have been equally
consistent. The transfer assumption survives its first test rather than
being confirmed by it, and it is still the only instrument bounding the
fourteen of sixteen arm~$\times$~model cells we do not submit. Two limits
survive: the flag is an upper bound, so true corrections are at or below
the table's, and transfer across \emph{models} remains untested, which is
where the estimator does most of its work.

\section{Harness Asymmetries, the Digest Defect, and the Substituted Primary
Model}
\label{app:harness}

\paragraph{A validator defect, \mGLM{} only.} \mGLM{}'s governed arms
carried a defect in the validator: a check intended to reject
\texttt{SELECT *} also rejected \texttt{COUNT(*)}, because the latter parses
with a \texttt{Star} node. The agent was told \emph{``SELECT * is not
allowed''} about queries containing no \texttt{SELECT *}: 161 rejections
across 124 of 401 tasks, and replaying the 160 distinct rejected queries
showed 79 were falsely rejected. A rejection whose stated reason is not true
of the query cannot be complied with, and the transcripts show the agent
guessing at unrelated clauses before finding a formulation that passed. This
handicapped both governed arms and neither baseline, so on this axis
\mGLM{}'s 55.1\% is a floor rather than a ceiling. \mDS{} ran with the fix,
and the contract arm's total count of queries rejected by the validator (218
on \mGLM{}, the 161 above among them) falls to 29, confirming the diagnosis.
The defect's direction is known; its magnitude is not separable from the
model change, and we do not attempt to estimate it.

\paragraph{Different endpoints across runs.} The runs used different
provider endpoints as well as different models, and \mSON{} a different kind
of route altogether: an enterprise gateway fronting Bedrock, on the Messages
API with cache control. \mSON{}'s dollar figures are that gateway's billed
rates under Messages-API cache pricing; on the same gateway's
OpenAI-compatible route the arms would have been billed at multiples that
differ by arm ($1.86{\times}$ on \armS{} against $3.40{\times}$ on \armM{},
replaying \mGPT{}'s rows at the same rates), so the \mSON{} cost ordering is
a property of the route as well as of the arms.

\paragraph{The row limit.} Symmetric result limits are not a detail. An
earlier iteration of this experiment used a 1{,}000-row limit, and the
governed arm looked $11{\times}$ more expensive than the baselines; the cause
was oversized result payloads poisoning the context window, not the
contract. The reported runs cap every result at 50 rows, but not perfectly
symmetrically: the cap reached the governed arms' \texttt{run\_query} but
not their \texttt{preview\_table}, which the library clamps at 100 rows of
its own accord. A governed arm could therefore see 100 rows from a preview
where an ungoverned arm saw 50 from \texttt{execute\_sql}. That happened on
31 preview calls across \mGLM{}, \mDS{} and \mGPT{}, against many thousands
of tool calls: small, but in the treatment's favour. \mSON{} ran with the
cap in place (484 preview calls, none over 50 rows); the other three are as
they ran.

\paragraph{The forced-answer policy.} When an agent exhausts its output
budget the harness forces a final answer instead of discarding the row. On
\mGLM{} that produced one additional correct answer out of 59, nearly
worthless for accuracy, but it eliminated unscoreable rows, and those rows
were 41 \armS{}, 18 \armM{} and zero in either governed arm. Dropping them
would have inflated the two baselines by silently discarding their hardest
tasks. In the arm order of the tables (\armS{}, \armH{}, \armM{}, \armC{}), \mDS{}
repeats the pattern (49, 15, 38 and zero on its complete-case set; 55, 15
and 42 before it), \mSON{} forces 43, 27, 32 and zero, and \mGPT{} forces 8
answers in \armH{} and none elsewhere. \armC{} is the only arm
that never exhausted its budget on any run.

\paragraph{The digest defect.} The contract digest is stamped
unconditionally, so \armH{}'s rows on \mGLM{}, \mDS{} and \mGPT{} carry the
\emph{real} contract's hash (\texttt{sha256:e438ecf7\ldots}) instead of the
hollow artifact's (\texttt{sha256:c46a767d\ldots}), pinned to a file that arm
never loaded. No result changes. Which artifact each arm loaded is fixed by
the harness, and the hollow arm's empty-placeholder responses and the
$n$-gram tests of Section~\ref{sec:hollow} evidence it independently. What is
missing is the per-row tamper-evidence we claim for \armH{}. The harness is
fixed; the released rows of the three earlier runs are left as they ran, and
\mSON{}, executed after the fix, stamps the hollow digest on all 401 of its
\armH{} rows.

\paragraph{The substituted primary model.} The pre-registration named
\texttt{deepseek-v4-pro-0813} as the primary comparison. Before sweeping it
we probed the bare-schema arm on 50 stratified tasks per candidate; on the
40 hard tasks all candidates scored, \mGLM{} answered 6, \mDS{} 10,
\texttt{deepseek-v4-pro} 11 and \mGPT{} 17. Pro against \mDS{} is three
discordant pairs (1 against 2), $p{=}1.0$; \mGPT{} against each of the
others is $p{\le}0.0156$. Pro would therefore not have supplied the
capability contrast the primary was chosen to provide, and \mGPT{} was
substituted by the same pre-specified criterion. What makes that a
substitution rather than a researcher degree of freedom is that the probe
measured the covariate, not the effect: model selection used bare-schema
accuracy while blind to the \armC{}-versus-\armM{} contrast the paper is
about. The public prediction of Section~\ref{sec:ceiling} was recorded
against \mGPT{} after the substitution was made. \mSON{} was added after a
three-model draft of the paper had been written; its rows carry the
commit, the contract digest and the hollow digest, which is the evidence
that it ran the frozen harness.

\section{Accuracy Inside and Outside the Contract's Domain}
\label{app:domain}

Table~\ref{tab:feebucket} is the split behind Section~\ref{sec:domain}.

\begin{table}[htbp]
\centering
\footnotesize
\setlength{\tabcolsep}{3pt}
\begin{tabular}{@{}llrrrr@{}}
\toprule
\tabhead{Model} & \tabhead{Bucket} & \armS{} & \armH{} & \armM{} & \armC{} \\
\midrule
\mGLM{} & fee (294)     & 13.9\% & 18.7\% & 24.5\% & \textbf{59.2\%} \\
\mGLM{} & non-fee (38)  & 5 & 9 & 4 & 9 \\
\mDS{} & fee (196)     & 23.0\% & 22.4\% & 44.9\% & \textbf{61.2\%} \\
\mDS{} & non-fee (30)  & 6 & 7 & \textbf{9} & 8 \\
\mSON{} & fee (294)     & 22.4\% & 40.1\% & 24.1\% & \textbf{73.5\%} \\
\mSON{} & non-fee (38)  & 10 & 8 & 8 & \textbf{11} \\
\mGPT{} & fee (294)     & 37.1\% & 53.4\% & 53.4\% & \textbf{83.7\%} \\
\mGPT{} & non-fee (38)  & \textbf{14} & 13 & 10 & 11 \\
\bottomrule
\end{tabular}
\caption{Hard-task accuracy split by whether the question concerns the
domain the contract describes. Non-fee cells are counts, not rates.}
\label{tab:feebucket}
\end{table}

Thirty of the 38 non-fee tasks are one family (\emph{most expensive MCC}, or
\emph{ACI}, \emph{for a transaction of $N$ euros}; 10 and 20 tasks), which
needs the fee rules without naming fees and so counts as \emph{derived} in
Section~\ref{sec:ceiling}. It scores 0--3 of 30 for \emph{every} arm on
\mGLM{}, \mDS{} and \mSON{}. \mGPT{} partially solves it (\armS{} 8, \armH{}
5, \armC{} 4, \armM{} 2 of 30), and that is where the bare baseline's lead on
the non-fee slice comes from. The 38 non-fee hard tasks are not all among
the 77 tasks with no fee semantics in Section~\ref{sec:ceiling}: this split
counts questions that name fees, that one counts questions that need them.

\section{The Margin Over a Prompt Is Not Monotone in Capability}
\label{app:interaction}

\begin{figure}[htbp]
\centering
\includegraphics[width=\columnwidth]{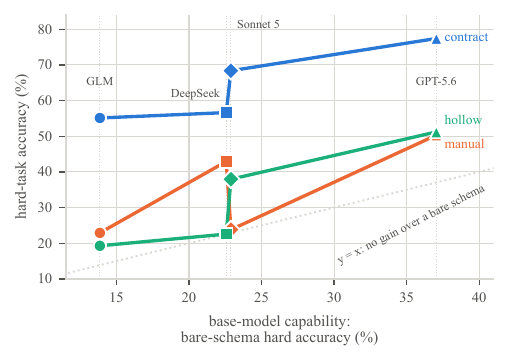}
\caption{Hard accuracy against base-model capability, where capability is
the bare-schema arm's own accuracy. The margin over \armM{} is $+32.2$,
$+13.7$, $+44.6$ and $+27.1$: not monotone.}
\label{fig:interaction}
\end{figure}

Ordered by what each model achieves from a bare schema
(Figure~\ref{fig:interaction}), the contract's margin over \armM{} on hard
tasks runs $+32.2$, $+13.7$, $+44.6$ and $+27.1$ on \mGLM{}, \mDS{},
\mSON{} and \mGPT{}. With the two flash models alone the pattern reads
as a clean interaction: \armM{} improved by 20.0 points while \armC{} moved 1.5,
and the lines converged. Two readings were available. Structured context
might substitute for capability, so that a stronger model extracts the same
semantics from prose unaided and the margin keeps shrinking; or \armC{}
might have been pinned near a benchmark ceiling at ${\approx}56\%$ while
\armM{} rose to meet it. \mGPT{} reached 77.4\%, so there is no ceiling at
56\%, and its $+27.1$ refutes the shrinking margin; \mSON{}'s $+44.6$ is the
largest of the four, on a model whose bare-schema score ties \mDS{}'s.
Whatever governs the size of the contract's advantage over a hand-written
prompt, it is not a monotone function of base capability, and four points
cannot support a third story. \mSON{} adds one thing: the bare-schema score
is a poor proxy for what a model does with structure. \mDS{} and \mSON{} are
0.3 points apart on it and 11.8 points apart with the contract.

\section{Reading a Leaderboard Number}
\label{app:leaderboard}

Section~\ref{sec:prior} quoted the two coupling costs the sources document
themselves: a tuned artifact is coupled to the serving model, and a Guide
tuned against a benchmark has to be measured on questions it has not seen.
The two systems atop DABStep's validated leaderboard make the point
concrete. The first reports 89.95\% hard with a lightweight inference
model driving a helper library distilled offline by a heavyweight one, built
by ``validating them against ground truth answers'' and by testing candidate
functions ``against the ground truth of multiple interconnected tasks'',
with pitfalls extracted ``from the test data'' and injected into the
inference prompt~\cite{nvidia-kgmon}. The second, at 87.57\%, publishes under
the title \emph{Beyond Fine-tuning: Solving DABstep's Hard Mode with
Versioned Assets}~\cite{oceanbase-datapilot}; we could not retrieve that post
in full and make no claim about its mechanism. Nothing in the benchmark
forbids using its labelled dev split this way, and we allege no violation;
where the first system says ``the test data'' we read its own loose usage
rather than an admission. Such numbers answer a different question from
ours: how much of a labelled, enumerable question space can be compiled away
in advance.

Our gold reconstruction supplies one datum on the same point. Across the
2{,}199 published DABStep submissions, organisations submitting once achieve
a median hard-split accuracy of 19.6\%, while those submitting 21 or more
times achieve 54.2\%. That gradient is equally consistent with genuine
iterative improvement. But on a small-template benchmark with a public
leaderboard, a single accuracy carries little information about
generalisation unless the submission count and the artifact's provenance
are stated alongside it. Ours are: one sweep per model, no tuning against
the task set, a digest-pinned artifact frozen 35 commits before the first
run, and two leaderboard submissions, one per arm of the main contrast on
one of the four models, both made after the runs were complete and neither
followed by a change to the artifact. They exist to move that contrast onto
official grading and to audit our own golds (\appref{app:golds}), not
to place.

The comparable artifacts are unpublished: MotherDuck's and NVIDIA's are
fused to the test set and coupled to the model they were tuned against, and
OceanBase's is described only in a post we could not retrieve. Ours is in
the repository with a digest because a contract written from a vendor
manual has neither coupling. That
is the sense in which a data contract is a different object from a semantic
layer, and it costs us the accuracy gap rather than excusing it.

Two claims that might look distinctive here belong to others. Paired testing
across several models is not distinctive~\cite{semantic-paired,dbt-benchmark},
and paired McNemar is standard practice for comparing learners on one
run~\cite{dietterich}. The contribution is the decomposition.

\section{The Two Escalations, in the Models' Own Words}
\label{app:escalations}

The post-run integrity check flagged two rows in 6{,}416: task~68, \armS{}
on \mDS{}, and task~2546, \armM{} on \mSON{}. Task~68 is the only one of the
24 attempting tasks on the two flash models to reach the database: DuckDB
\texttt{TEMP} tables die with the connection, and only this attempt
escalated to persistent tables.

The escalation was deliberate, and the model narrates it. It first issued
\texttt{CREATE TEMP TABLE feats}, and on the next call received
\texttt{Table with name feats does not exist!}. Its reasoning:

\begin{quote}\small\itshape
``TEMP table per session? The previous CREATE TEMP TABLE in the same
connection. The tool might use separate connections per call, so temp tables
don't persist. Let me create a regular table\ldots''
\end{quote}

and, having established that they now persist, \emph{``Since tables persist
across calls (non-temp), I can do sequential UPDATE operations.''} It then
hand-rolled Gaussian elimination across twelve statements (\texttt{DROP
TABLE IF EXISTS m1; CREATE TABLE m1 AS \ldots} through \texttt{m6}) to solve
a linear system, and answered the question wrongly as well.

\mSON{} reached the same place by a shorter route. On task~2770 it
lost a \texttt{TEMP} table between calls and reasoned \emph{``The temp tables
might be getting dropped between separate execute\_sql calls if they're
using different connections''}, the identical diagnosis, and batched the
chain into a single call, which kept it temporary. On task~2546 it went
straight to \texttt{CREATE OR REPLACE TABLE}, then \texttt{ALTER TABLE
\ldots ADD COLUMN} and \texttt{UPDATE} against its own persistent table, to
hold a merchant-month bucket join it could not get right as a subquery. Both
tasks were answered wrongly.

The statements are not malformed SQL that a validator would catch; the
ungoverned tool opens a fresh read-write connection per call and passes them
through, so \texttt{CREATE TABLE} succeeds silently. Because \texttt{TEMP}
tables are connection-scoped, a temp table never survives to the next tool
call, so the model receives no signal that it did anything wrong, only that
its own table vanished, and the natural repair is to stop using
\texttt{TEMP}. The run used a disposable per-worker copy of the warehouse,
so nothing real was harmed.

The idiom is a property of the model family as much as of its capability.
\mGPT{} uses common table expressions on 59\%/55\% of its ungoverned traces
(\armS{}/\armM{}), against 32\%/30\% on \mGLM{}, 26\%/27\% on \mDS{} and
32\%/37\% on \mSON{}. \texttt{CREATE TEMP} appears in 22 of \mGLM{}'s
ungoverned traces, 3 of \mDS{}'s, 8 of \mSON{}'s and none of \mGPT{}'s; on
\mGLM{} it accounts for every mutating statement submitted, on \mDS{} for 4
of 25. We cannot separate any of this from a trained disposition against
writing to a warehouse.

\fi

\end{document}